\documentclass[lettersize,journal]{IEEEtran}
\usepackage[mathlines,switch, pagewise]{lineno} 
\usepackage{amsmath,amsfonts}
\usepackage{array}
\usepackage[caption=false,font=normalsize,labelfont=sf,textfont=sf]{subfig}
\usepackage{textcomp}
\usepackage{stfloats}
\usepackage{hyperref}       
\usepackage{url}
\usepackage{verbatim}
\usepackage{graphicx}
\usepackage{cite}

\usepackage{inconsolata}

\usepackage{placeins} 
\usepackage{booktabs}       
\usepackage{amsfonts}       
\usepackage{nicefrac}       
\usepackage{microtype}      
\usepackage{lipsum}
\usepackage{fancyhdr} 
\usepackage{amsmath}
\usepackage{algorithm}
\usepackage[noend]{algorithmic}
\usepackage[bottom]{footmisc}

\begin{document}

\title{XiYan-SQL: A Novel Multi-Generator Framework For Text-to-SQL}

\author{Yifu Liu, Yin Zhu, Yingqi Gao, Zhiling Luo, Xiaoxia Li, Xiaorong Shi, Yuntao Hong, Jinyang Gao, Yu Li, Bolin Ding, Jingren Zhou,~\IEEEmembership{Fellow,~IEEE}
\thanks{
Yifu Liu, Yin Zhu, Yingqi Gao, Zhiling Luo, Xiaoxia Li, Xiaorong Shi, Yuntao Hong, Jinyang Gao, Yu Li, Bolin Ding, and Jingren Zhou are affiliated with Alibaba Cloud Computing Co., Ltd., Hangzhou, Zhejiang, China (emails: zhencang.lyf@alibaba-inc.com; sherlin.zy@alibaba-inc.com; gaoyingqi.gyq@alibaba-inc.com; godot.lzl@alibaba-inc.com; suian.lxx@alibaba-inc.com; shixiaorong.sxr@alibaba-inc.com; jiaxu.hyt@alibaba-inc.com; jinyang.gjy@alibaba-inc.com; lojze.ly@alibaba-inc.com; bolin.ding@alibaba-inc.com; jingren.zhou@alibaba-inc.com).

}}
\markboth{IEEE Transactions on Knowledge and Data Engineering,~Vol.~38, pp.~2474--2487, April~2026}
{XiYan-SQL: A Novel Multi-Generator Framework For Text-to-SQL}


\maketitle


\begin{abstract}
To leverage the advantages of LLM in addressing challenges in the Text-to-SQL task, we present XiYan-SQL, an innovative framework effectively generating and utilizing multiple SQL candidates. 
It consists of three components: 1) a Schema Filter module filtering and obtaining multiple relevant schemas; 2) a multi-generator ensemble approach generating multiple high-quality and diverse SQL queries; 3) a selection model with a candidate reorganization strategy implemented to obtain the optimal SQL query.
Specifically, for the multi-generator ensemble, we employ a multi-task fine-tuning strategy to enhance the capabilities of SQL generation models for the intrinsic alignment between SQL and text, and construct multiple generation models with distinct generation styles by fine-tuning across different SQL formats.
The experimental results and comprehensive analysis demonstrate the effectiveness and robustness of our framework. Overall, XiYan-SQL achieves a new SOTA performance of 75.63\% on the notable BIRD benchmark, surpassing all previous methods. It also attains SOTA performance on the Spider test set with an accuracy of 89.65\%.
\end{abstract}


\section{Introduction}
\IEEEPARstart{T}{he} task of translating natural language (NL) queries into structured query language (SQL) queries, known as Text-to-SQL, or NL2SQL, is a long-standing task for the development of NL interfaces to relational database management systems. 
This capability significantly reduces the interaction cost of users accessing database systems. 
Recent research is empowered by advances in large language models (LLMs), which have significantly improved the performance of NL2SQL solutions \cite{supersql}.
A series of approaches based on in-context learning (ICL) has been developed to enhance the SQL generation performance of these models by designing different prompts, such as chain-of-thought (CoT) \cite{10.5555/3600270.3602070}, question decomposition \cite{macsql}, and choice of demonstration examples \cite{mcssql}. 
Nonetheless, these methods are highly constrained by the sensitivity of LLMs to the structure and content of prompts.

Consequently, a commonly adopted strategy is to generate multiple SQL candidates by designing various forms of prompts, followed by an SQL selection process.
Compared to a single SQL query, this approach \cite{mcssql,macsql} of utilizing multiple forms of prompts enhances the diversity of the candidate samples and improves performance on challenging benchmarks such as BIRD \cite{bird}.
However, we observe that increased candidate diversity alone does not guarantee correctness or robustness: generated candidates can still contain errors and inconsistencies. Compact, task-specialized fine-tuned SQL models can often provide higher-quality candidates in this setting \cite{codes, dts_sql}. Moreover, utilizing self-consistency as a selection criterion does not always yield the optimal SQL query\cite{chasesql}.

To address these challenges, we propose XiYan-SQL\footnote{``XiYan" is the Chinese name of our ChatBI product.}, an end-to-end framework built on a novel architectural philosophy: a hybrid ensemble of expert generators. Instead of relying on a single, monolithic model, our approach is centered on a core of specialized, fine-tuned models, augmented by the ICL-based model. This sophisticated integration of multiple, diverse generators allows us to achieve superior accuracy and diversity.
We first introduce a Schema Filter Module that provides SQL generators with diverse and high-quality database schemas. 
This module performs multi-path retrieval of tables, columns, and values based on the inputs and subsequently employs a model for iterative column selection.

To realize this ``society of experts" \cite{minsky1986society} philosophy, we propose a novel generation method that integrates our diverse ensemble of generators. At its core are multiple fine-tuned models, for which we capitalize on the high controllability of supervised fine-tuning (SFT) to construct multiple SQL generation models.
Specifically, inspired by the multi-tasking approach in natural language processing, we explore a multi-task joint fine-tuning strategy tailored for SQL generation. 
This strategy involves the incorporation of various SQL and natural language transformation tasks to enhance the model's syntactic alignment capabilities. 
Additionally, we investigate a multi-format SQL enhancement training approach designed to improve diversity. 
This involves fine-tuning specialized models to preferentially generate semantically equivalent queries that differ in structure or style, for instance, using complex Common Table Expressions (CTEs) versus simpler subqueries, or enforcing specific stylistic conventions.
Ultimately, these strategies significantly improve the performance of fine-tuned models.

For the SQL candidate selection phase, we employ a selection model that works in concert with a candidate reorganization strategy. The strategy first reorganizes candidates based on consistency results to enhance the model’s attention towards potentially correct options. 
Subsequently, the selection model, fine-tuned on constructed comparative samples, identifies the best answer from this structured list, leading to improved selection performance.

In the experiments, we evaluate XiYan-SQL on two benchmarks, BIRD \cite{bird}, and Spider \cite{spider}, to demonstrate the superiority of the framework.
For the well-known and challenging BIRD benchmark, we achieve an execution accuracy of 75.63\%.
This result established a new SOTA performance on the BIRD leaderboard upon our submission on December 17, 2024.
For the Spider test set, we achieve SOTA performance with an execution accuracy of 89.65\%.
The extensive experimental results and comprehensive analysis further confirm the effectiveness and robustness of our method. 
The related code and models are being incrementally released through our project's main repository to support community research.
\footnote{https://github.com/XGenerationLab/XiYan-SQL}

Our main contributions are as follows.
\begin{itemize}
\item We propose XiYan-SQL, a novel Text-to-SQL framework that effectively integrates multiple SQL generators to generate high-quality and diverse SQL candidates, achieving new SOTA performance on both the BIRD and Spider benchmarks.
\item We introduce a Schema Filter Module that generates multiple high-quality schemas with different precision and recall through multi-path retrieval and iterative column selection.
\item We implement effective fine-tuning methods to construct multiple SQL generators, including enhancing the model's adaptability to task complexity through multi-task fine-tuning strategies, thus developing high-accuracy generation models, and constructing diverse models with different generation advantages through multi-format SQL enhancement training.
\item We propose a novel selection approach, combining a candidate reorganization strategy with a powerful SQL selection model to identify the optimal SQL from multiple candidates.
\item Comprehensive experimental results demonstrate the effectiveness of the XiYan-SQL framework and its key components.
\end{itemize}

This paper is organized as follows. 
In Section \ref{related_work}, we review the development of Text-to-SQL and the current work related to LLMs. 
In Section \ref{Methodology}, we introduce the complete XiYan-SQL framework, along with detailed explanations of the methods for Schema Filtering, Multiple SQL Generation, and SQL Selection. 
Then, in Section \ref{sec4}, we present a thorough and comprehensive experimental evaluation. 
Finally, we provide important empirical insights related to the XiYan-SQL technology in Section \ref{Discussion}, and conclude the paper in Section \ref{Conclusion}.

\section{Related Work}
\label{related_work}
In this section, we discuss the development of Text-to-SQL related work.
Text-to-SQL systems have garnered significant attention in recent years due to their potential to bridge the gap between natural language processing and SQL execution. 
Early work in this field relied mainly on hand-crafted templates \cite{10.5555/1864519.1864543}. 
With the advancement of transformers \cite{transformer} in deep neural networks, typical models have adopted an encoder-decoder architecture to generate SQL directly from natural language inputs. 
Several pre-trained language models (PLMs), such as BERT \cite{devlin2019bert}, TaBERT \cite{TaBERT}, and GraPPa \cite{yu2021grappa}, have been developed to encode structured data representations effectively.
However, these early methods demonstrate limitations when applied to more complex and diverse scenarios.
Currently, the rapid development of large language model technologies has demonstrated unique emerging capabilities in developing Text-to-SQL solutions. 
This includes advancements in prompt engineering techniques \cite{DAIL}\cite{dinsql}\cite{chasesql}, as well as pre-training or fine-tuning methods \cite{dts_sql}\cite{codes}\cite{yang2024synthesizing} for large language models.
In this paper, we focus mainly on the technology related to LLMs.

\noindent \textbf{Prompt Engineering Methods.}\hspace{1em}
Prompt engineering methods leverage the inherent capabilities of models to guide LLMs in generating diverse SQL queries by optimizing prompts \cite{dong2023c3, DAIL, chasesql, dinsql}. 
These methods are typically employed out-of-the-box on closed-source models with a vast number of parameters, such as GPT-4\cite{gpt4} and Gemini\cite{gemini15}.
Some approaches based on in-context learning address various stages of the workflow, including schema linking\cite{chess, distillery}, SQL generation \cite{supersql}, and SQL refinement\cite{dinsql, macsql, xie2024decomposition}, by designing question/task decomposition, demonstration selection\cite{mcssql}, chain-of-thought reasoning\cite{10.5555/3600270.3602070}, and multi-agent processing within the prompts \cite{macsql}.
Regarding schema linking, a common objective in prior work is to prune the database schema to a single, most-relevant subset. 
This includes modern LLM-based methods (e.g., CHESS \cite{chess}, CHASE-SQL \cite{chasesql}, OpenSearch-SQL \cite{opensearch}) and earlier sophisticated systems that use dedicated classifiers (e.g., RESDSQL \cite{RESDSQL}).
Even within the fine-tuning paradigm, methods like CodeS \cite{codes} tackle schema linking, albeit by integrating it into a monolithic model.
While their techniques vary, their goal is a single optimized schema for a single generator. In stark contrast, our Schema Filter's objective is to iteratively generate a diverse set of complementary schemas (e.g., a high-precision and a high-recall schema). This diversity is purpose-built to provide robust inputs for our downstream multiple SQL generation module.
\\
To further enhance performance, an effective strategy is to generate multiple candidate SQL queries through different prompt designs and subsequently select the best one \cite{mcssql, chasesql}.
However, these methods rely solely on prompt design, which presents challenges related to the sensitivity of LLMs to prompts, potentially compromising the quality and robustness of SQL generation, along with incurring significant inference overhead.
In contrast, our approach effectively mitigates these problems by constructing different high-accuracy fine-tuned SQL generation models.

\noindent \textbf{Fine-tuning Methods.}\hspace{1em}
The fine-tuning methods can align objectives in models with fewer parameters, allowing the fine-tuned models to demonstrate improved controllability in generating SQL queries \cite{codes, dts_sql, momq2024, yang2024synthesizing, tkde10930803}.
Additionally, this method offers lower inference overhead and enhanced data privacy protection.
Recently, CodeS\cite{codes} has been trained on a large corpus related to SQL, aimed at improving the general Text-to-SQL performance of smaller models. 
DTS-SQL\cite{dts_sql} introduces a two-stage supervised fine-tuning method that decomposes the task into two simpler components: schema linking and SQL generation. 
Yang et al. \cite{yang2024synthesizing} constructed the SENSE model by fine-tuning it on synthesized robust data and applying Direct Preference Optimization \cite{dpo} on weak data.
Despite these advancements, these methods exhibit limited performance and often struggle in more challenging scenarios or benchmarks.

While sharing high-level goals with the works discussed, XiYan-SQL distinguishes itself through its unique architectural philosophy and its relationship with the emerging agent paradigm.
Architecturally, unlike prompt-based methods that query a single API model multiple times, or prior fine-tuning works that focus on improving a single model, we are the first to propose and systematically evaluate a hybrid framework centered on an ensemble of distinct, fine-tuned local models. 
This approach is not merely an assembly of parts, but a synergistic co-design: our iterative schema filter is built to produce diverse schemas for the multiple generators, and our novel multi-task/multi-format fine-tuning strategy is specifically created to induce the expert ``personalities" required for the ensemble's success.
In the context of agent-based systems, our framework's fully decoupled and modular design allows its components to be seamlessly embedded as powerful ``tools" or ``experts." 
For instance, a general-purpose agent could delegate complex database interaction tasks to our entire pipeline. Thus, XiYan-SQL is positioned not as an alternative to the agent paradigm, but as a complementary and enabling technology for it.
By demonstrating the power of this integrated, multi-generator architecture, our work presents a novel and practical blueprint for building robust, private, and cost-effective Text-to-SQL systems that can both stand alone and enhance future agent-based workflows.

\begin{figure*}[!tp]
    \centering
    \includegraphics[width=1.0\textwidth]{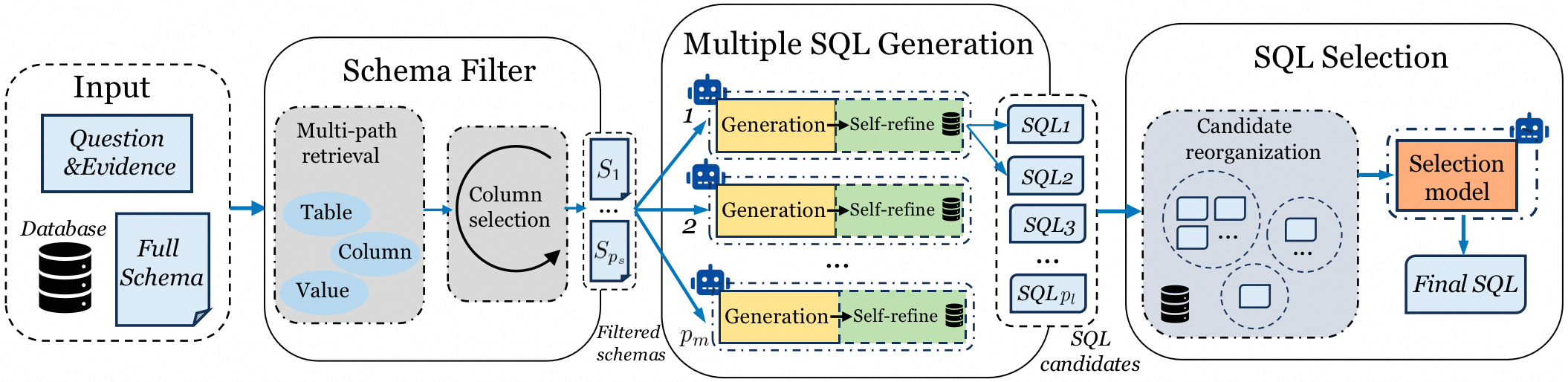}
    \caption{Overview of the proposed XiYan-SQL framework, including three steps: Schema Filter, Multiple SQL Generation, and SQL Selection.}
    \label{fig:pipeline}
\end{figure*}

\section{Methodology}
\label{Methodology}
\subsection{Overall Framework}
We present the XiYan-SQL framework in Figure \ref{fig:pipeline}. 
The inputs are: (i) the user question; (ii) external textual knowledge (hereafter ``evidence"), provided in the form of short text snippets such as schema explanations, synonym mappings, or business rules; and (iii) the database schema (tables, columns, and values).
The framework consists of three steps: Schema Filter, Multiple SQL Generation, and SQL Selection.
The Schema Filter module initially generates multiple filtered schemas.
Subsequently, multiple candidate SQL queries are generated through various SQL generators. 
Finally, the most accurate SQL query is selected by the selection model based on the candidate reorganization. 
The details of each step are described below.

\subsection{Schema Filter}
Our Schema Filter Module consists of multi-path retrieval and column selection, designed to effectively implement schema linking for large databases and generate multiple filtered schemas.
\\
\textbf{Multi-path Retrieval.}\hspace{1em}Multi-path Retrieval is a pruning mechanism for large databases that employs cosine similarity for the embeddings of all tables, columns, and values.
Initially, we used an out-of-the-box LLM $\mathcal{M}_s$ to extract keywords from the question $Q$ and evidence $E$, represented as $K = f_{\mathcal{M}_s}(Q,E)$.
These keywords, denoted as $K = \{ k_1, k_2, \ldots \}$, are instrumental in identifying relevant columns and values within the database.
In the table and column retrieval stage, we compute the scores between the keywords $K$ and the original schema $S$, which can be represented as a set of columns, $\{ c_1, c_2, \ldots \}$.
This process is calculated as the product of two distinct scores as follows.

\begin{equation}
\resizebox{0.4\textwidth}{!}{$\text{Score}(k_i, c_j) = \left\langle \mathbf{V}_{Q||E}, \mathbf{V}_{Tab({c_j})} \right\rangle \cdot \left\langle \mathbf{V}_{k_i}, \mathbf{V}_{c_j} \right\rangle$}
\label{eq:col_retriever_score}
\end{equation}
One part is the cosine similarity between the embedding of the keyword $\mathbf{V}_{k_i}$ and that of the column metadata $\mathbf{V}_{c_j}$.
The other part is the calculation between the input text and the table metadata.
In detail, we concatenate the query and evidence to obtain the embedding representation $\mathbf{V}_{Q||E}$.
The table to which $c_j$ belongs can be determined by projection $Tab$, and the embedding representation of the table is $\mathbf{V}_{Tab({c_j})}$.
Through this process, we identify the top-$k$ columns that are most relevant to each keyword.

For the value retrieval, we first use edit distance to locate the top-$k$ values of each column that are similar to each keyword.
Subsequently, we employ the RoBERTa tokenizer to uniquely tokenize the value text along with column metadata. 
This step enhances the efficiency of Locality Sensitive Hashing \cite{datar2004locality} in filtering text related to values. 
An embedding cosine similarity is then applied, using a threshold to refine the retrieved values. 
Through multi-path retriever, we filter out a schema $S^{rtrv} = \{ c_{r_1}, c_{r_2}, \ldots \}$ from the original schema $S$.
\\
\textbf{Column Selection.}\hspace{1em}Column Selection introduces diversity into the retrieved schemas, facilitating subsequent multiple SQL generation.
This process is detailed in Algorithm \ref{alg:schema selector}, which outputs $p_s$ different schemas through multiple iterations.
In each iteration, we prompt an LLM $\mathcal{M}_s$ to select columns related to the question and evidence, saved as $S^{slct}_i$. 
The function $\text{PFKeyIdentifier}$ then identifies primary and foreign keys according to the columns in $S^{slct}_i$.
It is designed to handle multi-column keys by leveraging the database's metadata: it automatically completes partial keys (i.e., including all columns of a composite key if one is selected) and includes the corresponding referenced keys from other tables to preserve valid join paths.
A new schema is generated by unifying the previous schemas with the selected columns.
We update $S^{rtrv}$ by removing only the selected non-key columns and explicitly retaining key columns (primary/foreign keys) to maintain schema integrity and connectivity.
As iterations increase, the precision of the selected schemas tends to decrease while recall improves, resulting in a diversity of filtered schemas.
\renewcommand{\algorithmicrequire}{ \textbf{Input:}}
\renewcommand{\algorithmicensure}{ \textbf{Output:}}
\begin{algorithm}
\caption{Column Selection Algorithm}
\label{alg:schema selector}
\begin{algorithmic}[1]
    \REQUIRE Set of columns $S^{rtrv} = \{ c_{r_1}, c_{r_2}, \ldots \}$, Question $Q$, Evidence $E$, The maximum iteration $p_{s}$
    \ENSURE Schema set $\mathcal{S} = \{ S_1, \ldots, S_{p_s}\}$
    \STATE Initialize a list $\mathcal{S} \gets [\ ]$
    \FOR{$i = 1$ to $p_{s}$}
        \STATE $S^{slct}_i \gets f_{\mathcal{M}_s}(S^{rtrv}, Q, E)$  \COMMENT {Select from $S^{rtrv}$}
        \STATE $P_i \gets \text{PFKeyIdentifier}(S^{slct}_i)$ 
        \STATE $S_i \gets \left( \bigcup_{k=1}^{i-1} S_k \right) \cup S^{slct}_i \cup P_i$ \COMMENT {Unify and generate $S_i$}
        \STATE Append $S_i$ to $\mathcal{S}$
        
        \STATE $S^{rtrv} \gets S^{rtrv} \setminus (S^{slct}_i \setminus P_i)$ \COMMENT{Remove selected non-key columns, preserve PK/FK.}
        
    \ENDFOR
    \RETURN $\mathcal{S}$
\end{algorithmic}
\end{algorithm}

\vspace{-5pt}

\subsection{Multiple SQL Generation}
\label{sec3_3}
Generating multiple candidate SQL queries to explore a broader search space to improve performance has been demonstrated as a reasonable approach in previous studies \cite{chasesql, mcssql}. 
However, prompt-based methods are constrained by the model's sensitivity to prompt formats, presenting a challenge in ensuring SQL quality while enhancing diversity. 
Considering that supervised fine-tuning can better align the model with preferred behaviors \cite{zhou2024lima, tunstall2024zephyr}, we explore a dedicated training strategy to develop SQL generation models that generate high-quality and diverse candidates.

\begin{figure}
    \centering
    \includegraphics[width=0.48\textwidth]{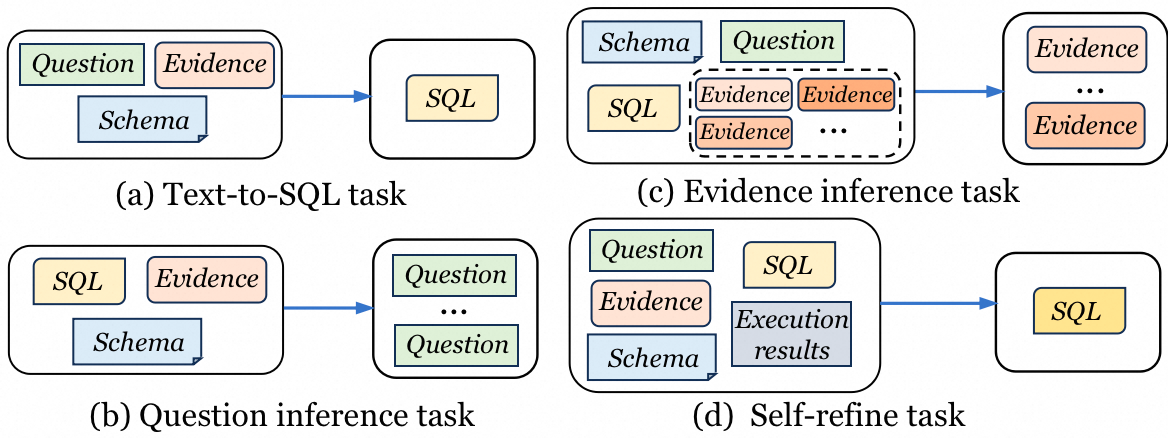}
    \caption{Illustration of the four tasks in our Multi-Task Fine-Tuning strategy. It shows the input-output transformations for (a) the standard Text-to-SQL task, and the auxiliary tasks of (b) Question Inference, (c) Evidence Inference, and (d) Self-Refine.}
    \label{fig:mt1}
\end{figure}

As a translation task, Text-to-SQL aims to achieve semantic alignment between structured representations and natural language \cite{liu2024survey}. 
Drawing inspiration from previous multi-task approaches \cite{multitask_better, multitask_better2} that leverage auxiliary tasks for learning, we explore a multi-task fine-tuning approach that utilizes related tasks to enhance information alignment.
The standard task process for converting text to SQL, given the inputs of the question, database schema, and evidence, is illustrated in Figure \ref{fig:mt1}(a).

However, this unidirectional training (Text-to-SQL) is insufficient to build a deep, bidirectional understanding between language and SQL.
To address this issue and enhance the semantic alignment capability of LLMs for this task, we further design auxiliary alignment tasks and explicitly fine-tune the SQL generation model for execution.
In response to natural language questions, we design a reverse inference task, as illustrated in Figure \ref{fig:mt1}(b), which aims to encourage the model to infer a set of potential questions based on SQL and relevant information.
For natural language evidence knowledge, we establish a task focused on the reverse inference of evidence, shown in Figure \ref{fig:mt1}(c). 
This task involves the model identifying the most relevant evidence from a set of candidate evidence based on the SQL generation process.
Additionally, for SQL queries, we develop a self-refine task for the model, as depicted in Figure \ref{fig:mt1}(d), which entails regenerating SQL after refinement based on previous SQL generation processes and execution results.
By leveraging a series of specialized tasks to jointly fine-tune models, our developed single generator surpasses ICL-based closed-source models, such as GPT-4o and Gemini, resulting in the generation of higher-quality SQL queries.
The detailed data construction methods for these auxiliary tasks are provided in Appendix II.B to ensure reproducibility.

\begin{figure}
    \centering
    \includegraphics[width=0.5\textwidth]{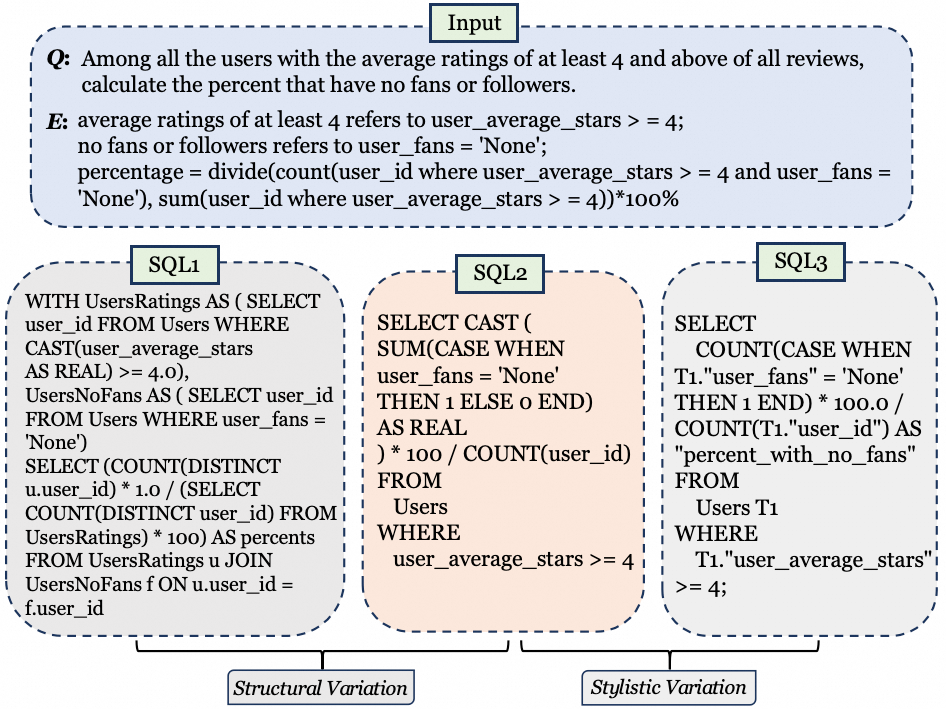}
    \caption{Examples of ``multi-format" SQL queries corresponding to the same input. Compared to the typical response (middle), the left query demonstrates a Structural Variation (e.g., using a more complex structure), while the right query illustrates a Stylistic Variation (e.g., adopting a different writing convention).}
    \label{fig:ft1}
\end{figure}

We further develop from a single generator to multiple generators to enhance the diversity of generated candidates. 
For a single user question, there can be multiple SQL queries that correspond to fulfill the intent.
Therefore, we aim to develop additional models with different ways of generation to increase the likelihood of arriving at the correct answers through diverse outputs.
To achieve this, we introduce ``multi-format" SQL variations during model construction. These added reformulations fall into two primary types: 
(1) Structural Variations: Encouraging the model to use more complex structures (e.g., rewriting a subquery as a Common Table Expression (CTE)), as illustrated by $\mathrm{SQL}_1$ in Figure \ref{fig:ft1}.
(2) Stylistic Variations: Adopting different writing conventions (e.g., standardizing keyword casing and table aliases), as shown by $\mathrm{SQL}_3$ in Figure \ref{fig:ft1}.
These multi-format data are then used to train specialized generators, with the construction methods for these diverse variations detailed in Appendix II.B.
By leveraging these formatted and multi-task data for joint fine-tuning, we obtain high-accuracy models with diverse SQL generation preferences.

As shown in Figure \ref{fig:pipeline}, we provide the generation model with $p_s$ different schemas generated during the Schema Filter phase to expand the candidate pool. 
To further enhance the diversity of candidates, our framework not only integrates multiple fine-tuned models but also incorporates an ICL-based model to collectively generate $p_l$ candidate SQL queries, where $p_l = p_s * p_m$, and $p_m$ represents the number of generators. 
Moreover, for each generation, if the execution of the generated SQL encounters syntax errors or anomalous values, the corresponding generator will utilize execution feedback to regenerate a SQL query through its inherent self-refine capabilities.
The complete process for multiple SQL generation is illustrated in Algorithm \ref{alg:sql_gen}. 
Please refer to Appendix I for details about the ICL-based generator.

\renewcommand{\algorithmicrequire}{ \textbf{Input:}} 
\renewcommand{\algorithmicensure}{ \textbf{Output:}} 

\begin{algorithm}
\caption{Multiple SQL Generation Algorithm}
\label{alg:sql_gen}

\begin{algorithmic}[1]
    \REQUIRE Question $Q$, Evidence $E$, Filtered Schema set $\mathcal{S} = \{ S_1, \ldots, S_{p_s}\}$, List of SQL Generators $\mathcal{M}$, DataBase $D$
    \ENSURE Candidate SQL list $L = \{ l_1, \ldots, l_{p_l}\}$
    
    \STATE Initialize Candidate SQL $L \gets \emptyset$
    \FOR{$i = 1$ to $p_{s}$}
        \FOR{$j = 1$ to $p_{m}$}
            \STATE Predict SQL $l_{ij} \gets f_{\mathcal{M}_j}(Q, E, S_i)$
            \STATE Execute Result $r \gets \text{execute}(l, D)$
            
            \IF{No exception in $r$}
                \STATE Append $l_{ij}$ to $L$
            \ELSE
                \FOR{retry $1$ times}
                    \STATE $\hat{l}_{ij} \gets f_{\mathcal{M}_j}(Q, E, S_i, l, r)$ \COMMENT{Self-refine}
                    \STATE Append $\hat{l}_{ij}$ to $L$
                \ENDFOR
            \ENDIF
        \ENDFOR
    \ENDFOR
    \RETURN $L$
\end{algorithmic}
\end{algorithm}

\subsection{SQL Selection}
After obtaining multiple SQL candidates, the challenge remains to select the most accurate and reasonable SQL query from these candidates.
Most candidate selection methods \cite{mcssql,chess} employ self-consistency \cite{wang2023selfconsistency} to select the SQL query that appears most consistently across multiple candidate samples.
However, there are limitations present: it cannot adequately handle situations where there is a lack of majority consensus, and even the most consistent result is not always the correct case.
To this end, we propose a model-based selection approach, which consists of a selection model and a candidate reorganization strategy to guide its attention.

To develop an effective selection model, we focus on constructing a rich listwise comparison dataset. 
This process begins with generating diverse candidate SQL queries using multiple generators, which include our ensemble of fine-tuned models and powerful ICL-based models (e.g., GPT-4o).
With these candidates, we construct training instances composed of positive and negative samples.
Recognizing that multiple queries can be semantically correct, we identify positive samples not just by matching the ground-truth SQL, but by verifying any candidate that produces an identical execution result. 
Negative samples are obtained through a multi-pronged strategy, involving not only natural failures (e.g., syntax, execution, or logical errors) but also more challenging ``near-miss" examples created via LLM-based ``controlled modifications"—a process where we introduce subtle, deliberate errors into correct queries. 
Before final dataset assembly, all candidates undergo a ``de-formalization process": we apply a standard SQL formatter to remove superficial stylistic differences (e.g., casing and spacing), forcing the model to learn semantic correctness rather than spurious generator-specific patterns. 
A comprehensive breakdown of this entire data construction pipeline is provided in Appendix II.C to ensure reproducibility.

During inference, to help the model better handle the numerous candidates, we apply the aforementioned candidate reorganization strategy to enhance its attention.
To align with the training phase, all candidates presented to the selection model are first de-formalized to remove superficial stylistic artifacts.
The complete selection process is illustrated in Algorithm \ref{alg:selector}. 
Initially, we obtain the execution results $R = \{r_1, \ldots, r_n\}$ of the candidate SQL queries $L = \{l_1, \ldots, l_n\}$ on the database, ensuring that any candidates with execution errors have already been excluded. 
We group the candidates based on the consistency of their execution results to obtain a collection of clusters $\mathcal{C}=\{C_1,..., C_m\}$. 
Subsequently, we perform inter-group and intra-group sorting. 
The inter-group sorting is performed in descending order based on the size of each group, resulting in $\mathcal{C}'$. 
The intra-group sorting is executed according to the order of the generators $\mathcal{O}$ corresponding to the candidates, 
resulting in $\mathcal{C}''$, where the sequence $\mathcal{O}$ is arranged in descending order based on the performance evaluation of the generators.
\renewcommand{\algorithmicrequire}{ \textbf{Input:}} 
\renewcommand{\algorithmicensure}{ \textbf{Output:}} 
\begin{algorithm}
\caption{SQL Selection Algorithm}
\label{alg:selector}
\begin{algorithmic}[1]
    \REQUIRE  Candidate SQL queries $L$, Question $Q$, Evidence $E$, Filtered Schema set $\mathcal{S} = \{S_1, \ldots, \}$, Selection Model $\mathcal{M}_c$, SQL Execution results $R$, Generators Order $\mathcal{O}$
    \ENSURE Selected SQL $l^*$
    \STATE $L \gets \text{deformalize}(L)$ \COMMENT{Ensure consistency}
    \STATE Clustering $\mathcal{C}=\{C_1 ,..., C_m\}\gets \text{groupby}(L,R)$
    \STATE Schema $\mathcal{S}^{un}\gets \text{schema\_union}(\mathcal{S})$
    
    \IF{$|\mathcal{C}| = 1$} 
        \RETURN $l^*\gets \arg_{l\in L}\min |l|$ \COMMENT{All are consistent; select shortest as a heuristic for simplicity.}
    \ELSE
        \STATE Clusters $\mathcal{C}' \gets \text{sort}(\mathcal{C})$ by size, descending \COMMENT {Inter-group sorting}
        \STATE Initialize Clusters $\mathcal{C}'' \gets \emptyset$
        
        \FOR{each cluster $C'_i \in \mathcal{C}'$} 
            \STATE Append $C''_i \gets \text{sort}(C'_i)$ order by $\mathcal{O}$ \COMMENT {Intra-group sorting}
        \ENDFOR
        
        \STATE Initialize Candidate $L' \gets \emptyset$, indicator $j\gets0$
        \IF{$|C_1''|\geq \lceil{|L|}/2\rceil$}
            \FOR{each Cluster $C_i''\in \mathcal{C}''$}
                \FOR{each SQL $l \in C_i''$}
                    \STATE Append $L_j'\gets l$ , $j\gets j+1$
                \ENDFOR
            \ENDFOR
        \ELSE
            \FOR{each Cluster $C_i''\in \mathcal{C}''$}
                \STATE Append $L_j'\gets \arg_{l\in C_i''} \min |l|$, $j\gets j+1$
            \ENDFOR
        \ENDIF
        \RETURN $l^*\gets f_{\mathcal{M}_c}(Q, \mathcal{S}^{un}, E, L')$ \COMMENT{Selection model prediction}
    \ENDIF

\end{algorithmic}
\end{algorithm}

If there exists a consensus result that predominates within the clusters, namely $|C_1''|\geq \lceil{|L|}/2\rceil$, 
we organize all candidates sequentially according to order $\mathcal{C}''$ and present them to the selection model $\mathcal{M}_c$.
Conversely, we select one candidate from each group within $\mathcal{C}''$ and assemble the results in accordance with the inter-group ordering to present to $\mathcal{M}_c$.
The model $\mathcal{M}_c$ takes into account the question, the union of all filtered schemas, the evidence, and the reorganized candidates to make its decision. 
Due to the selection tendency of LLMs towards options that appear earlier \cite{zheng2024large}, employing this clustering and ranking approach to organize candidates allows us to effectively leverage strong consistency priors while mitigating the influence of interfering information. 
This allows the model's attention to be concentrated, thereby enhancing its comprehension ability.

\section{Experiments}
\label{sec4}
In this section, we aim to provide a detailed evaluation of XiYan-SQL. 
We first introduce the experimental setting, followed by the demonstration of XiYan-SQL's SOTA performance on important benchmarks. 
Subsequently, we conduct ablation experiments on the overall framework and provide a comprehensive analysis of each component.
To ensure fairness in our experiments, the results of previous methods used for comparative analysis have been sourced from publicly available benchmark tests, which typically represent their most powerful implementations. 
Additionally, the results we have reproduced for comprehensive comparison are obtained using the same configuration as XiYan-SQL (e.g., base models, etc.).

\subsection{Experimental Setup}

\noindent \textbf{Dataset.}\hspace{1em}
We evaluate the XiYan-SQL framework on two of the most recognized and challenging benchmarks in modern Text-to-SQL research: BIRD and Spider.

\noindent BIRD \cite{bird}, focusing on complex real-world databases, is the most challenging large-scale cross-domain Text-to-SQL benchmark.
It contains 12,751 unique question-SQL pairs and 95 big databases with a total size of 33.4 GB.
The training, development, and unpublished test set contain 9,428, 1,534, and 1,789 pairs, respectively.
We use its development set for local evaluation and submit our performance results on the unreleased test set for a fair evaluation.
Additionally, we utilize the provided Oracle knowledge as evidence.

\noindent Spider \cite{spider}, widely used for Text-to-SQL evaluation, is a large-scale, complex, cross-domain benchmark.
It comprises 10,181 questions and 5,693 distinct SQL queries across 200 databases, covering 138 different domains.
This dataset is divided into training, development, and test sets, comprising 8,659, 1,034, and 2,147 examples, respectively, with a primary focus on evaluating the results on the test set.

\noindent \textbf{Metrics.}\hspace{1em}
Following the previous studies, we use Execution Accuracy (EX) to evaluate the performance of Text-to-SQL methods.
Since an SQL query can be expressed in various forms, EX is used to assess the validity of the predicted SQL query and to determine whether the execution results are consistent with the ground-truth SQL query.
It should be noted that there are slight differences in the calculation of the Execution Accuracy between the BIRD and Spider benchmarks; however, their overall objective remains consistent.

In addition to EX, the latest version of the BIRD benchmark introduces a Reward-based Valid Efficiency Score (R-VES) to evaluate the execution efficiency of correctly predicted SQL queries. 
R-VES is an adjusted version of the previous VES, which assesses the model by considering both the accuracy of and the runtime performance of SQL queries \cite{bird}. 
We also present an evaluation of this metric on the BIRD benchmark in the main results.

\noindent \textbf{Implementations.}\hspace{1em}
In the proposed framework, all database schemas are represented in the form of the M-schema \cite{xiyansql}.
For the Schema Filter, we utilize GPT-4o \footnote{gpt-4o-0806 for all GPT-4o (https://openai.com/index/gpt-4/)} \cite{openai2023gpt4} as the $\mathcal{M}_s$ model for keyword extraction and column selection.
We set the maximum number of iterations $p_s$ to 2 to obtain the filtered schema set $\mathcal{S} = \{ S_1, S_2\}$.
For the Multiple SQL Generation part, we first leverage GPT-4o and Qwen-Max \footnote{https://qwenlm.github.io/zh/blog/qwen2.5-max/} \cite{qwen25} for assistance in constructing relevant multi-task and multi-format SQL data. 
We conduct experiments on multiple models of varying sizes to thoroughly demonstrate the superiority of our fine-tuning approach. 
To achieve top performance with XiYan-SQL, we select the widely used Qwen2.5-Coder-32B \cite{qwen25coder} as the foundation for constructing the final multiple fine-tuned SQL generators. 
We utilize a total of $p_m=5$ generators. As detailed in Section \ref{sql_generators}, these consist of four specialized fine-tuned models—a base multi-task model, models for structural and stylistic variations, and a mixed-data model—and one ICL-based generator ($\mathrm{SQLG_5}$), ultimately resulting in the generation of $p_l = 10$ SQL candidates.
Due to limited cost and resources, we have not increased the number of models or candidates.
For the SQL selection part, we also utilize GPT-4o to augment the training data. 
Given that the selection task is relatively simpler than the generation task, we choose Qwen2.5-Coder-7B for developing the SQL selection model.
All base models used for fine-tuning in the following experiments are their respective instruction versions.
For more implementation details, please refer to Appendix II.

\begin{table}[!tp]
\caption{Performance comparison of different Text-to-SQL methods on BIRD benchmark.}
\label{tab:bird_main}
    \centering
    \scalebox{1.0}{
    \begin{tabular}{llll}
    \toprule
    Methods & EX(Dev) & EX(Test) & R-VES(\%)\\     
    \midrule

    GPT-4o & 58.47 & - & 51.75 \\
    TA-SQL + GPT-4\cite{tasql} & 56.19  & 59.14 &  - \\
    DAIL-SQL\cite{DAIL} & 54.76 & 57.41 &  54.02\\

    SFT CodeS-15B \cite{codes} & 58.47 & 60.37 & 61.37 \\
    SuperSQL \cite{supersql} & 58.50 & 62.66 & - \\ 
    MCS-SQL \cite{mcssql} & 63.36 & 65.45 & 61.23 \\ 
    PURPLE + RED + GPT-4o & 68.12 & 70.21 & 65.62 \\
    
    Insights AI & 72.16 & 70.26 & 66.39 \\ 
    CHESS \cite{chess}& 68.31  & 71.10 & 66.53\\ 
    Distillery + GPT-4o \cite{distillery}& 67.21 & 71.83 & 67.41 \\ 
    OpenSearch-SQL, v2 + GPT-4o & 69.30 & 72.28 & 69.36\\ 
    ExSL + granite-34b-code & 72.43 & 73.17 & 71.37 \\ 
    DSAIR + GPT-4o & 74.32 & 74.12 & 70.13 \\ 
    CHASE-SQL+Gemini\cite{chasesql}& 74.46 & 74.79 & 70.57 \\ 
    \midrule
    XiYan-SQL & 73.34 & \textbf{75.63} & \textbf{71.41} \\ 
    \bottomrule
    \end{tabular}
    }

\end{table}

\subsection{Main Results}

\noindent \textbf{BIRD Results.}
We present the end-to-end performance of the XiYan-SQL on the BIRD benchmark, based on the official benchmark page.
In addition to the results obtained from GPT-4o and XiYan-SQL, all other results are derived from the respective original publications.
As shown in Table \ref{tab:bird_main}, our method achieves an EX score of 75.63\%. This performance established a new SOTA on the highly competitive BIRD leaderboard at the time of submission (Dec 17, 2024), surpassing all previous works and outperforming the second-place entry by 0.84\%.
This provides strong evidence for the effectiveness of our ``society of experts" philosophy, validating our integrated approach to candidate generation and selection.

We further demonstrate the comparison of XiYan-SQL against other methods based on the R-VES metric, with results sourced from the official benchmark webpage. 
As shown in Table \ref{tab:bird_main}, we achieve an R-VES score of 71.41\%, maintaining our position as SOTA performance on the BIRD leaderboard, comparable to EX.
\\
\textbf{Spider Results.}
To further evaluate the generalizability of XiYan-SQL, we conduct experiments on the Spider test set, with results shown in Table \ref{tab:spider_main}. 
Similar to the results obtained on the BIRD benchmark, XiYan-SQL achieves an EX score of 89.65\%, surpassing previous methods and establishing a new SOTA performance. 
This result, achieved without any Spider-specific tuning, demonstrates the framework's strong performance consistency. It shows that our approach is effective not only on the value-heavy complexity of BIRD but also on the structural complexity emphasized by Spider, complementing our primary findings.

\begin{table}[!tp]
\caption{Performance comparison of different Text-to-SQL methods on Spider test benchmark.}
\label{tab:spider_main}
\centering
\scalebox{1.0}{
\begin{tabular}{p{6.0cm}c}
 \toprule
Methods & EX(\%) \\     
\midrule
MCS-SQL + GPT-4~\cite{mcssql} & 89.60 \\ 
CHASE-SQL + Gemini 1.5~\cite{chasesql} & 87.60 \\ 
PET-SQL~\cite{petsql} &  87.60 \\ 
SuperSQL~\cite{supersql} & 87.00 \\
DAIL-SQL + GPT-4~\cite{DAIL} & 86.60 \\ 
DPG-SQL + GPT-4 & 85.60 \\ 
Tool-SQL + GPT-4~\cite{tool_sql} & 85.60 \\
DIN-SQL + GPT-4~\cite{dinsql} & 85.30 \\
GPT-4o & 83.54\\ 
C3 + ChatGPT + Zero-Shot~\cite{dong2023c3} & 82.30 \\ 
\midrule
XiYan-SQL & \textbf{89.65} \\ 
\bottomrule
\label{tab:results_spider_test}
\end{tabular}
}

\end{table}

\subsection{Ablation Study}
To investigate the significance of each component within the full pipeline, we conduct ablation studies to assess the incremental impact of each component on the EX.
The results of the end-to-end ablation on the BIRD dev set are presented in Table \ref{tab:ablation}.
To begin with, when the Schema Filter module is removed, leaving only the full schema, it significantly affects the quality of the generated SQL candidates, leading to an overall performance drop of approximately 1.24\%.
This result highlights that the proposed schema filter method is crucial for generating high-quality schemas, thereby enhancing the performance of the SQL generators.
When we perform an ablation on Multiple SQL Generation by utilizing a single fine-tuned generation model to generate one SQL query, we observe a significant performance decline of approximately 4\%. 
This result strongly underscores the outstanding performance of our multiple SQL generation technique. 
Additionally, if the single generation model is not derived from the advanced training strategies we propose, but rather from a conventional fine-tuned model or a model based on ICL, the performance degradation would be even more pronounced. 
We will further discuss this in the next section.
Finally, when we remove the proposed SQL Selection approach, which relies solely on the consistency of the candidate SQL execution results for a majority voting, we observe a significant decrease in performance. 
This strongly corroborates the importance of this specific technique within the overall framework.
Overall, our ablation experiments indicate that each component of XiYan-SQL plays a vital role in achieving high accuracy. 

\begin{table}[!tp]
\caption{Ablation studies of each component on the performance of XiYan-SQL on BIRD dev.}
\label{tab:ablation}
    \centering
    \begin{tabular}{p{4.5cm}cc}
    \toprule
    Methods &  EX(\%) & $\Delta$EX(\%) \\    
    \midrule
    XiYan-SQL (Full pipeline) & 73.34 & -\\ 
    \midrule
    \textbf{w/o} Schema Filter & 72.10 &-1.24 \\
    \textbf{w} Only one SFT generator & 69.30 & -4.04 \\
    \textbf{w} Only majority voting   & 70.21 & -3.13 \\
    \bottomrule
    \end{tabular}

\end{table}

\subsection{Study of Schema Filter}
We first evaluate the performance of our Schema Filter module in filtering database schemas.
Our module iteratively generates a set of diverse schemas. 
In our experiments, we use the first two schemas generated, denoted as $S_1$ (from the first iteration) and $S_2$ (from the second iteration). 
Results for these, alongside different schema filter methods on BIRD, are shown in Table \ref{tab:schema_filter}.
We employ the evaluation metrics presented in schema linking \cite{distillery}, where $P$ represents the column's precision in the filtered schema, while $R_c$ and $R_v$ denote the recall for columns and values, respectively.
Compared to the advanced methods for schema linking that we reproduce, our Schema Filter achieves higher recall for both columns and values while maintaining precision. 
Furthermore, as iterations increase, it effectively adjusts the precision and recall of the schema, resulting in diverse schemas.

\begin{table}
\caption{Schema Filter comparison of different Text-to-SQL methods on BIRD dev. Our Schema Filter (schema $S_1$) and (schema $S_2$) refer to the schemas generated in the first and second iterations of our filter, respectively, showcasing the trade-off between precision and recall.}
\label{tab:schema_filter}
    \centering
    \begin{tabular}{p{4.0cm}cccc}
    \toprule
    Methods & $P$(\%) & $R_c$(\%) & $R_v$(\%) \\     
    \midrule
    Full schema & 10.14 & 100 & -  \\
    CHESS~\cite{chess} & 85.90 & 74.77 & 63.71\\ 
    TA-SQL~\cite{tasql}& 82.50 & 76.83 & - \\
    \midrule
    Our Schema Filter (schema $S_1$) & 74.89 & 83.64 & 91.31  \\
    Our Schema Filter (schema $S_2$) & 54.90 & 89.77 & 93.63  \\
    \bottomrule
    \end{tabular}

\end{table}

We further supplement the performance of the final generated SQL queries under different schema linking methods.
Due to performance differences among the various generators, we average the results from three different models: the fine-tuned model based on Qwen2.5-Coder-32B, GPT-4o, and Gemini-1.5-pro \footnote{https://deepmind.google/technologies/gemini/}. 
Notably, we emphasize the $\Delta$EX metrics, which indicate the improvements achieved compared to the full schema method.
As shown in Table \ref{tab:schema_filter_append}, our methods demonstrate significant enhancements. 
Additionally, we find that selecting a single iteration (i.e., $p_s=1$) yields better results than the second iteration. 
Throughout the entire pipeline, we introduce greater schema diversity through multiple iterations.
For additional experiments about the Schema Filter, please refer to Appendix III.

\begin{table}
\caption{The comparison of the end results of different Schema Filter methods on Bird dev.}
\label{tab:schema_filter_append}
    \centering
    \begin{tabular}{p{4.5cm}cccc}
    \toprule
    Methods & EX(\%)  & $\Delta$EX(\%)\\     
    \midrule
    Full schema & 61.57 & - \\
    CHESS~\cite{chess} & 63.30 & +1.73 \\ 
    \midrule
    Our Schema Filter (schema $S_1$) & \textbf{63.75} &+2.18 \\
    Our Schema Filter (schema $S_2$) & 62.97 &+1.40 \\
    \bottomrule
    \end{tabular}

\end{table}

\subsection{Study of Single SQL Generation}
\label{sec:E_single_gen}
We first investigate the performance of a single fine-tuned generator in generating a single SQL query based on the proposed fine-tuning strategy. 
Table \ref{tab:MSG} demonstrates the consistency improvements achieved through the multi-task fine-tuning strategy across different model sizes, and we also compare these results with those of directly using GPT-4o as an SQL generator.
For the ICL-based methods, by incorporating effective demonstration examples, we observe an approximate 4\% improvement compared to directly using the powerful GPT-4o. 
Furthermore, regardless of the model type or size, the SQL generation models based on our multi-task training strategy exhibit significant enhancements, with improvements ranging from 8.3\% to 15.3\%.
Compared to conventional fine-tuning, our joint multi-task training approach significantly enhances the model's understanding of the translation from natural language to SQL, thereby improving the model's adaptability to task complexity. 
This is reflected in the performance improvements across five widely selected basic fine-tuned models, with increases of 4.5\%, 4.04\%, 2.9\%, 2.9\%, and 2.2\%, respectively.
The fine-tuned performance of our models with sizes of 14B and above surpasses that of the few-shot approach using GPT-4o. When we employ Qwen2.5-Coder-32B as the base model, the fine-tuned SQL generation performance can reach 66.88\%, which is also the SOTA performance for single SQL models.

\begin{table}[!tp]
\caption{Comparison of EX of different single generators on BIRD dev. All evaluation methods exclude additional steps such as schema linking and self-refine. Note that the base models we used for fine-tuning are all their instruct versions.}
\label{tab:MSG}
    \centering
    \begin{tabular}{p{6.0cm}c}
    \toprule
    Methods &  EX (\%) \\    
    \midrule
    GPT-4o + zero-shot & 58.47 \\ 
    GPT-4o + few-shot  & 62.65 \\
    \midrule
    Base model (\textit{DeepSeek-Coder-V2-Lite}) & 40.61 \\
    SFT generator & 51.49 \\     
    SFT generator + multi-task & \textbf{55.99} \\
    \midrule
    Base model (\textit{Qwen2.5-Coder-7B}) & 50.39 \\
    SFT generator & 56.06 \\     
    SFT generator + multi-task & \textbf{60.10} \\
    \midrule
    Base model (\textit{Qwen2.5-Coder-14B}) & 55.28 \\
    SFT generator & 61.02 \\     
    SFT generator + multi-task & \textbf{63.95} \\
    \midrule
    Base model (\textit{Codestral-22B}) & 50.52 \\
    SFT generator & 62.32 \\     
    SFT generator + multi-task & \textbf{65.25} \\
    \midrule
    Base model (\textit{Qwen2.5-Coder-32B}) & 58.60 \\
    SFT generator & 64.67\\
    SFT generator + multi-task & \textbf{66.88} \\
    \bottomrule
    \end{tabular}
\end{table}

We further present a ``leave-one-out" ablation study of the various auxiliary tasks during the training phase. 
Table \ref{tab:task_ablation} reports the ablation results on two models to explore the impact of each auxiliary task, where No.1$\sim$No.5 represent the fine-tuning results based on the Codestral-22B model, and No.6$\sim$No.10 represents the fine-tuning results based on the Qwen2.5-Code-32B model.
The results clearly demonstrate the effectiveness and complementarity of our multi-task design. 
Removing the Question Inference task leads to the most significant performance drop, highlighting its critical role in semantic alignment. 
The removal of the Self-refine and Evidence Inference tasks also results in noticeable performance degradation, confirming their respective contributions to the model's robustness and reasoning capabilities. Crucially, the full multi-task model (No.6) consistently and significantly outperforms all ``leave-one-out" variants, as well as the baseline model trained only on the primary Text-to-SQL task (No.10), which validates our complete fine-tuning strategy.
It is also worth noting that the self-refine capability, learned during this multi-task training, can be selectively applied during the inference phase. 
As detailed in our methodology (Section \ref{sec3_3}), this inference-time refinement process can provide an additional performance boost of approximately 1\%, further enhancing the quality of the final candidate pool.

\begin{table}[!tp]
\caption{
The ablation results on the EX performance of multiple auxiliary tasks on BIRD dev.
}
\label{tab:task_ablation}
    \centering
    \scalebox{0.95}{
    \begin{tabular}{c|m{1.5cm}|m{1.5cm}|m{1.7cm}|c}
    \toprule
    No. &   \parbox{2cm}{Question inference task} & \parbox{2cm}{Evidence inference task} & \parbox{2cm}{Self-refine task} & EX(\%)  \\    
    \midrule
    1 & \centering \checkmark & \centering \checkmark & \centering\checkmark & 65.25\\ 
    2 &  & \centering\checkmark & \centering\checkmark & 63.75\\ 
    3 &\centering\checkmark  &  &\centering\checkmark  & 64.60\\ 
    4 &\centering\checkmark  &\centering\checkmark  &  & 64.08\\ 
    5 &  &  &  & 62.23\\ 
    \midrule
    6 & \centering \checkmark & \centering \checkmark & \centering \checkmark & 66.88\\ 
    7 &  &\centering \checkmark  &\centering \checkmark  & 65.91\\ 
    8 &\centering \checkmark  &  &\centering \checkmark  & 66.49\\ 
    9 &\centering \checkmark  &\centering \checkmark  &  & 65.71\\ 
    10 &  &  &  & 64.67\\ 
    \bottomrule
    \end{tabular}
    }

\end{table}

\subsection{Study of Multiple SQL Generation}
\begin{figure}[!tp]
\setlength{\abovecaptionskip}{0.1cm}  
\setlength{\belowcaptionskip}{-0.4cm}
    \centering
    \includegraphics[width=0.45\textwidth]{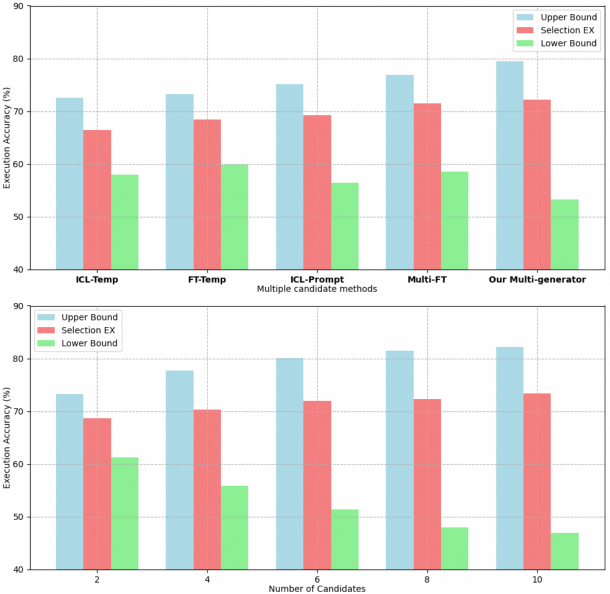}
    \caption{
    (a) Comparison of EX among different multiple candidate methods with five candidates (Figure Above).
    (b) Performance of multi-generator method under different candidate numbers (Figure Below).
    }
    \label{fig:expr1}
\end{figure}

We continue to evaluate the performance of utilizing multiple SQL generators to obtain multiple candidate SQL queries. Figure \ref{fig:expr1} (a) presents the comparison of various candidate generation methods to highlight the superiority of our multiple fine-tuned models and multi-generator ensemble approach. 
Here, we follow the calculations of upper-bound and lower-bound performance as described in \cite{chasesql}.
The ICL-Temp indicates the results obtained using GPT-4o with different sampling temperatures, which, as anticipated, exhibit lower candidate diversity and accuracy in the selected outputs.
The FT-Temp refers to the candidates generated by our fine-tuned generator with different sampling temperatures, which demonstrate higher accuracy than ICL-Temp but similarly lack good diversity.
The ICL-Prompt represents candidates generated by GPT-4o using different example selection prompts. 
Compared to the first two methods, ICL-Prompt shows improvements in both upper-bound and selection accuracy; however, the model's sensitivity to the prompts results in a lower-bound performance. 
The Multi-FT refers to the candidates generated from our multiple fine-tuned models with distinct generation advantages.
The fine-tuned generation models employed in this method achieve high accuracy, and the generated SQL queries also demonstrate considerable diversity, effectively balancing the relationship between lower and upper bounds. 
This results in an approximate 2\% accuracy improvement over ICL-Prompt. 
Furthermore, the best method we explored (i.e., the last one) incorporates an ICL-based generation method alongside our multiple fine-tuned models. 
This integration allows us to maintain high-quality generation while further enhancing diversity.

Figure \ref{fig:expr1} (b) illustrates the performance of our multi-generator ensemble approach as a function of the number of candidates. 
As these candidates are sampled from our pool of distinct generators, this analysis serves as a proxy for evaluating the impact of the number of generators.
The results for different numbers of candidates are obtained through multiple random samplings from the candidate pool of five SQL generators.
As the number of candidates increases, the candidate space exhibits greater diversity, leading to a higher upper-bound; at $p_l=10$, this upper-bound reaches 82.2\%, showing significant potential. 
However, the lower-bound also decreases considerably, which indicates that the selection process becomes crucial. 
Therefore, we explore an enhanced SQL selection method tailored for scenarios involving multiple candidates.
For a more direct and comprehensive analysis of the interaction between the number of generators and the Schema Filter iterations, please refer to the detailed study in Appendix IV.A.

In summary, the experiments in this section demonstrate that a multi-generator approach significantly outperforms single-generator, as well as simple sampling or multi-prompt methods. 
The success of this paradigm relies on both the quantity and, crucially, the diversity of the generators. While the multi-task strategy (Section \ref{sec:E_single_gen}) enhances the quality of each individual generator, it is our multi-format training strategy that is specifically designed to induce the structural and stylistic diversity across the ensemble.
To rigorously quantify the individual and synergistic contributions of these two strategies, a comprehensive, decoupled ablation study is provided in Appendix IV.B. 
The results therein confirm that the quality-enhancing multi-task strategy and the diversity-inducing multi-format strategy provide complementary benefits, validating our overall fine-tuning methodology.

\subsection{Study of SQL Selection}
After obtaining multiple candidate SQL queries, we introduce a strategy that combines candidate reorganization with a SQL selection model to choose the optimal candidate SQL. 
Next, we explore the selection strategy with a detailed comparative analysis. 
Table \ref{tab:sele_method} presents the analysis of different SQL selection methods. 
We observe that the majority voting strategy, based on consistency, achieves the EX of 69.82\%.
Furthermore, leveraging our fine-tuned selection model, we compare several distinct selection strategies. 
Here, $\mathrm{s}_1$ denotes presenting the candidate samples to the selection model in a random order, $\mathrm{s}_2$ indicates the organization of candidate SQL samples in descending order based on their corresponding model performance, and $\mathrm{s}_3$ employs the proposed candidate reorganization to enhance the model's attention.
Our selection strategy (\textit{i.e.}, $\mathrm{s}_3$) outperforms the majority voting method by approximately 3\% and demonstrates significantly better performance than the other selection strategies.
As the number of candidates increases, the selection task becomes more complex, and the advantages of our current method become even more pronounced. 
\begin{table}[!tp]
 \caption{Comparison of EX of different SQL Selection methods on BIRD dev.}
\label{tab:sele_method}
    \centering
    \begin{tabular}{p{4.0cm}cc}
    \toprule
    Methods &  $p_l=5$(\%) & $p_l=10$(\%) \\
    \midrule
    Majority voting & 69.94 & 70.21\\ 
    Our selection model (+ $\mathrm{s}_1$) & 69.69 & 68.19\\ 
    Our selection model (+ $\mathrm{s}_2$) & 71.45 & 71.84\\ 
    \midrule
    Our selection model (+ $\mathrm{s}_3$) & \textbf{72.29} & \textbf{73.34}\\ 
    \bottomrule
    \end{tabular}

\end{table}

In the aforementioned experiments, we have demonstrated that our selection model, based on the proposed candidate reorganization strategy, achieves performance that significantly surpasses that of the Majority Voting method.
This improvement highlights the powerful synergy between our selection strategy and the strong discriminative capability of our model. The strategy effectively amplifies the model's ability to focus on the correct SQL.
We further evaluate the performance of the selection model.
To minimize the impact of candidate quantity and order on the model, we randomly sample three different candidate quantities (2, 6, and 10) from the BIRD dev dataset and organize the candidates in random order for presentation to the model. 
As shown in Table \ref{tab:selection_model}, our fine-tuned selection model outperforms the baseline model and also exceeds the performance of advanced out-of-the-box models (such as GPT-4o and Gemini). 
Additionally, due to the smaller size of our fine-tuned model, it exhibits faster response times and improved instruction-following capabilities.

\begin{table}[!tp]
\caption{Evaluation of the independent SQL selection judgment ability of different selection models.}
\label{tab:selection_model}
    \centering
    \begin{tabular}{p{6.0cm}c}
    \toprule
    Methods & EX(\%) \\
    \midrule
    Qwen2.5-Coder-7B (Base) & 62.84 \\ 
    Gemini-1.5-pro & 66.30 \\
    GPT-4o & 67.47 \\
    Our fine-tuned selection model & \textbf{69.56}\\ 
    \bottomrule
    \end{tabular}

\end{table}

\subsection{Analysis of Results from Different SQL Generators}
\label{sql_generators}
In this section, we present the results of different generators that contribute to the final performance of XiYan-SQL. 
As previously mentioned, we ultimately utilize five distinct generators, with each generator generating two candidates corresponding to two schemas. 
We provide the individual evaluation results in Table \ref{tab:generator_res}, along with their contributions to overall performance.
Specifically, the roles of generators are as follows: the generator $\mathrm{SQLG_1}$ is a base fine-tuned model based on multi-task training;
$\mathrm{SQLG_2}$ is a specialized model fine-tuned on a dataset enriched with structural variations (e.g., CTEs);
$\mathrm{SQLG_3}$ is a specialized model fine-tuned on a dataset featuring specific stylistic variations;
$\mathrm{SQLG_4}$ is a comprehensive model fine-tuned on a mix of all multi-task and multi-format data;
$\mathrm{SQLG_5}$ is an ICL-based generator using the GPT-4o model.

Avg. EX represents the average execution accuracy of the two candidates from a single generator.
The CR (Contribution Ratio) is defined to measure the contribution of each generator in more difficult cases that require careful selection. 
The calculation is as follows: we first identify and exclude the ``unanimous" cases (approx. 45\% of the dataset) where all SQL candidates yield identical execution results. 
For the remaining cases, our SQL selection method chooses a single optimal candidate. 
The CR for a generator is the percentage of these non-unanimous cases where the selected candidate is generated by that generator. 
Since each selected candidate belongs to a single generator, the sum of all CRs is 100\%.

As observed, the generator $\mathrm{SQLG_1}$ exhibits the best individual performance (Avg. EX) and contributes the most significantly (CR), representing the strong capability of our multi-task fine-tuning strategy. 
Notably, $\mathrm{SQLG_2}$, which is trained to favor complex ``structural variations", exhibits a lower Avg. EX score than $\mathrm{SQLG_1}$. 
This suggests that generating these advanced forms is inherently harder and more error-prone.
A similar principle of trade-offs applies to $\mathrm{SQLG_3}$ and $\mathrm{SQLG_4}$; their respective focuses on strict stylistic consistency and broad solution-space coverage come at the cost of a minor drop in peak accuracy on standard patterns.
However, its non-trivial Contribution Ratio and its unique correct generations (shown by the $R_1$ value in Table \ref{tab:generator_dis}) demonstrate that it is essential for solving certain challenging queries that simpler forms cannot handle. 
These findings validate our core hypothesis: no single generation form is universally optimal; instead, integrating diverse generation pathways is a highly effective strategy for tackling a wide spectrum of real-world challenges.

The analysis of the result distribution in Table \ref{tab:generator_dis} further supports this, where the $R_n$ values denote the situation in which only $n$ generators have generated correct outputs, as explained in the table note.
As the model with the highest accuracy, $\mathrm{SQLG_1}$ exhibits high values of $R_3$ and $R_4$, suggesting that its generation is more robust and overall quality is higher. 
In contrast, the distributions of $\mathrm{SQLG_3}$ and $\mathrm{SQLG_4}$ are more concentrated within the middle range, exploring diverse generations while maintaining quality. 
Meanwhile, $\mathrm{SQLG_2}$ and $\mathrm{SQLG_5}$ exhibit higher $R_1$ values, indicating greater diversity; each possesses a unique ability to solve certain problems alone.
This illustrates the trade-off between quality and diversity inherent in our ensemble approach.

\begin{figure}[!tp]
    \centering
    \includegraphics[width=0.47\textwidth]{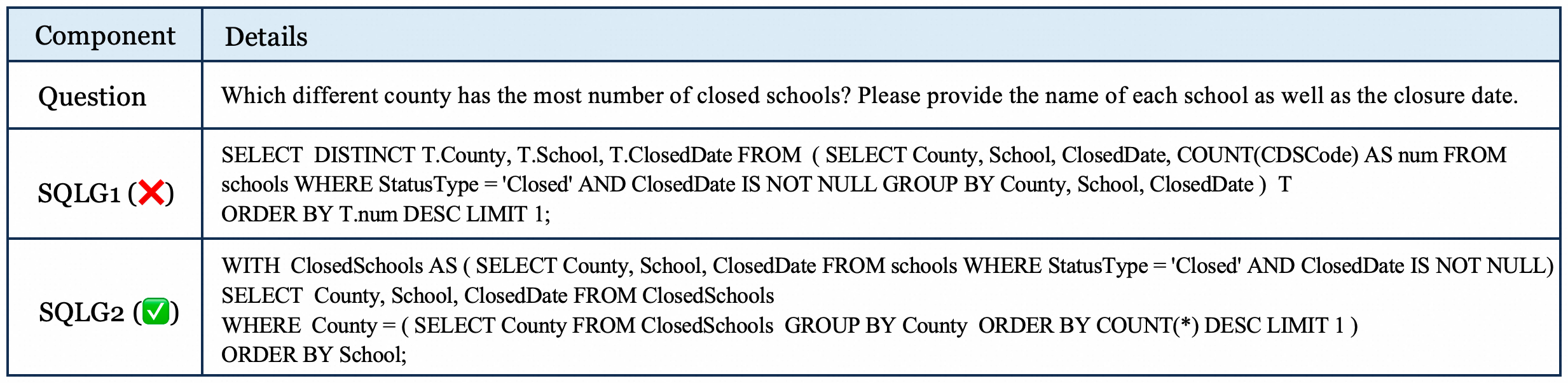}
    \caption{Case study demonstrating the value of Structural Variation on a challenging query from the BIRD dev. Inputs such as evidence and the schema are omitted for clarity.}
    \label{fig:case_level}
\end{figure}

\noindent \textbf{Case-Level Analysis}
We present a detailed case-level analysis from the BIRD dev set as qualitative evidence for the value of structural variation. 
The full details are shown in Figure \ref{fig:case_level}.
This question requires a two-step reasoning process: first, identifying the county with the most closed schools, and second, listing all schools within that specific county.
Our best-performing single model, $\mathrm{SQLG_1}$, failed to solve this. 
It generated a query with a flawed aggregation strategy, attempting to solve both reasoning steps at once. 
In contrast, the specialized model $\mathrm{SQLG_2}$ successfully resolved the query by generating a correct and more readable solution using a CTE, which effectively decomposed the problem into two distinct logical steps.
This case powerfully illustrates our core argument. 
The structural diversity introduced by $\mathrm{SQLG_2}$—specifically, its ability to generate a CTE—is not merely for variety.
It is a critical component for achieving correctness on challenging queries that require sophisticated logical decomposition.
While $\mathrm{SQLG_2}$ is not entirely generalizable, it achieves important complementarity with other models.

In summary, each generator plays an important and distinct role, and their performances align with our design objectives for the multi-generator framework of XiYan-SQL.

\begin{table}[!tp]
\caption{
Performance of the five individual generators on the BIRD dev set. 
These results are directly derived from the final pipeline and maintain all configurations.}
\label{tab:generator_res}
    \centering
    \begin{tabular}{m{2.5cm}m{1.5cm}m{1.0cm}}
    \toprule
    SQL Generators & Avg. EX(\%) & CR(\%) \\
    \midrule
     $\mathrm{SQLG_1}$ &  \centering 69.34\arraybackslash &  \centering 51.90 \arraybackslash \\
     $\mathrm{SQLG_2}$ &  \centering 66.05 \arraybackslash &  \centering 5.27 \arraybackslash\\
     $\mathrm{SQLG_3}$ &  \centering68.15\arraybackslash &  \centering6.13\arraybackslash\\
     $\mathrm{SQLG_4}$ &  \centering68.50\arraybackslash &  \centering30.55\arraybackslash \\
     $\mathrm{SQLG_5}$ &  \centering64.51 \arraybackslash &  \centering6.13 \arraybackslash\\
    \bottomrule
    \end{tabular}
\end{table}

\begin{table}[!tp]
\caption{
Distribution of correct results across different levels of inter-generator agreement on the BIRD dev. The column $R_n$ shows the percentage of correct answers from a generator in cases where exactly $n$ models succeeded. Thus, $R_1$ measures the contribution of unique solutions, while  $R_5$ represents universally correct answers.}

\label{tab:generator_dis}
    \centering
    \scalebox{1.0}{
    \begin{tabular}{lccccc}
    \toprule
    SQL Generators  & $R_1$(\%) & $R_2$(\%) & $R_3$(\%) & $R_4$(\%) & $R_5$(\%)  \\
    \midrule
       $\mathrm{SQLG_1}$   & 1.42 & 2.31 & \textbf{3.73} & \textbf{13.24} & 76.44 \\
       $\mathrm{SQLG_2}$   & \textbf{1.96} & 2.13 & 3.29 & 10.76 & 76.44 \\
       $\mathrm{SQLG_3}$   & 1.33 & \textbf{3.47} & 3.64 & 12.27 & 76.44 \\
       $\mathrm{SQLG_4}$   & 1.60 & \textbf{3.47} & 3.11 & 12.62 & 76.44 \\
       $\mathrm{SQLG_5}$   & \textbf{2.22} & 3.20 & 2.58 & 9.78 & 76.44 \\
    \bottomrule
    \end{tabular}}

\end{table}

\subsection{Analysis of Cost Efficiency}
The main advantage of the XiYan-SQL framework lies in its unique multi-generator ensemble approach, which enables the generation and utilization of multiple candidate SQL queries, thereby achieving optimal performance. 
However, this may raise concerns about the need to access the models multiple times and the associated inference latency when processing a single user query. 
To address this issue, we conduct a comprehensive discussion on the operational costs of XiYan-SQL.

During the training phase, fine-tuning the Qwen2.5-Coder-32B to obtain a SQL generation model requires approximately 45 GPU hours (using NVIDIA A100 80G GPUs), consuming a total of about 180 million tokens. 
Fine-tuning the Qwen2.5-Coder-7B to obtain a selection model takes about 15 GPU hours, with a total of approximately 50 million tokens consumed. 
Detailed information about the training phase costs is reported in Appendix II.

During the inference stage, Table \ref{tab:costs} summarizes the inference costs of XiYan-SQL. 
It is important to note that the above results reflect the average per-sample statistics of XiYan-SQL on the BIRD dev set. 
The SQL generation process may include certain self-refine operations, which have been accounted for. 
We utilize the official API to call GPT-4o, while all other fine-tuned models are deployed using minimal resources.
Additionally, due to factors such as network conditions and deployment environments, the results may exhibit some normal fluctuations.
In the Schema Filter module, most of the time cost during the retrieval phase arises from the model calls for keyword extraction and the multi-path similarity retrieval, while the time costs during the selection phase result from two iterations of column selection.
During the multiple SQL generation phase, the main time consumption involves generating multiple candidate SQL queries. 
The SQL Selection stage includes two parts: executing all candidates against the database and inferring the selection model to select the final result. 
On our test environment (Apple M4 Pro, 48GB RAM), the average execution time per sample (with 10 candidates) on BIRD dev is 0.35 seconds. 
It is important to note that this execution time is highly dependent on the specific hardware and the scale of the database. 
Ultimately, the total end-to-end latency per sample is approximately 40.81 seconds.

To further illustrate the inference cost, we compare XiYan-SQL with the previous SOTA method, CHASE-SQL\cite{chasesql}. 
The performance achieved by our method with 10 generated SQL candidates surpasses that of CHASE-SQL, which generates 21 SQL candidates. 
In this configuration, based on the inference information provided by CHASE-SQL, the overall latency of our pipeline is only 60\% of that of CHASE-SQL.

\begin{table}[!tp]

\caption{The average inference costs of the XiYan-SQL framework on BIRD dev.
*SQL execution latency is measured on a local machine (Apple M4 Pro, 48 GB RAM) and can vary depending on the hardware and database size.}
\label{tab:costs}
    \centering
    \scalebox{0.9}{
    \begin{tabular}{lccc}
    \toprule
    Stage  & Base model & \parbox{1.5cm}{Input/Output (tokens)} & \parbox{1.2cm}{Inference latency(s)}  \\
    \midrule
       Schema Filter (Retrieval)   & GPT-4o & 413.9/37 & 11.75 \\
       Schema Filter (Selection)   & GPT-4o & 3791.5/63.1 & 6.77  \\
       \parbox{3.2cm}{Multiple SQL Generation \\ (per candidate, avg.)}  & Qwen2.5-Coder-32B & 582.5/47.8 & 2.25   \\
       \parbox{3.2cm}{Multiple SQL Generation \\ (per candidate, avg.)}    & GPT-4o & 763.2/54.7 & 2.75   \\
       Candidate SQL Execution   & N/A (Database) & N/A & $0.35^{*}$   \\
       SQL Selection   & Qwen2.5-Coder-7B & 1052.3/1 & 0.78   \\
       All pipeline   & Multiple models & $\sim\!6700/\sim\!205$ & $\sim\!40.81$  \\
    \bottomrule
    \end{tabular}}
\end{table}

\subsection{Study of Generalization}
A key indicator of a Text-to-SQL framework's practical utility is its ability to generalize to diverse, unseen data and environments. We demonstrate the robust generalization of XiYan-SQL from two primary perspectives, both grounded in public, verifiable benchmarks.

First, we examine its in-distribution generalization within the BIRD benchmark. As presented in Table \ref{tab:bird_main}, XiYan-SQL achieves a 2.3\% higher execution accuracy on the black-box test set compared to the public dev set. This improvement margin is notably larger than that of previous state-of-the-art methods, which typically remain below 1\%. This result suggests that our framework learns generalizable translation patterns rather than merely overfitting to the development set.

Second, to provide compelling evidence of out-of-domain generalization, we evaluated XiYan-SQL on the BIRD-CRITIC benchmark \cite{li2025swe}. BIRD-CRITIC is a novel public benchmark, jointly developed by the BIRD team and Google Cloud, designed to diagnose and resolve complex user challenges in real-world applications. 
It represents a forward-looking trend in the community, moving beyond simple query translation to address practical, error-prone scenarios. We chose BIRD-CRITIC because it serves as a more rigorous and meaningful litmus test for a model's practical applicability than older datasets. 
As verified by the official public leaderboard \footnote{https://bird-critic.github.io/} (as of May 2025), XiYan-SQL achieved a score of 41.0\% on this challenging benchmark, reaching SOTA performance.
This outstanding result strongly validates that our approach is effective and adaptable in complex, realistic settings.
For a detailed description of the experimental setup, the specific adaptations made for the benchmark, and a snapshot of the leaderboard, please refer to Appendix VI.

In summary, the combination of strong performance on the BIRD test set and SOTA results on the public BIRD-CRITIC benchmark confirms that XiYan-SQL is not only a high-performing but also a highly generalizable framework, ready for real-world application challenges.

\section{Discussion}
\label{Discussion}
As a comprehensive Text-to-SQL solution, XiYan-SQL offers several empirical insights based on our experiments, presented as follows:
\begin{itemize}
\item For schema linking, it is crucial to balance precision and recall. By ensuring that recall reaches a certain threshold (e.g., 80\%), improving precision becomes significantly more beneficial for the overall result. This is evidenced by the various iterations of our Schema Filter.
\item For SQL generators, incorporating relevant multiple tasks can significantly improve the performance of fine-tuned models. 
These tasks may be adjusted based on thematic considerations. 
Furthermore, regarding multiple candidate approaches, whether through multiple generators or a single generator with multi-path generation, it is crucial to prioritize the quality of individual generation before increasing diversity in outputs.

\item In contrast, API-based approaches that leverage powerful reasoning models (e.g., o4-mini) exhibit superior generalization on certain SQL-reasoning tasks—for example, the ``SQL debug" task on the BIRD-CRITIC leaderboard. 
Our single fine-tuned model, however, demonstrates competitive performance on standard NL2SQL tasks such as BIRD and Spider.
To further improve results, we apply the full ensemble pipeline of XiYan-SQL, which enables us to outperform even these strong reasoning models. 
As detailed in Appendix V, a direct comparison on two complex internal datasets shows that while top-tier API models can surpass our standalone fine-tuned model, our full XiYan-SQL pipeline still achieves the highest accuracy, empirically validating our ensemble framework's effectiveness.

\item For SQL selection, similar to previous studies \cite{liu2024lost, zheng2024large}, we have also found that LLMs, whether fine-tuned or not, exhibit sensitivity to the order and organization of candidates. 
Therefore, supplementing with additional prior knowledge can help enhance performance.
\end{itemize}

\section{Conclusion}
\label{Conclusion}
This study presents a novel Text-to-SQL framework, XiYan-SQL, which focuses on generating high-quality and diverse candidate SQL queries, effectively selecting the best among them.
Specifically, we introduce a Schema Filter module that employs multi-path retrieval and iterative selection to obtain multiple high-quality schemas.
Then, we leverage the proposed multi-task and multi-format fine-tuning strategy to develop a variety of SQL generators, each with distinct advantages, enabling the generation of high-quality and diverse SQL candidates, which we demonstrate is crucial for solving complex queries where a single generation style may fail.
Finally, we propose an integrated selection module where a candidate reorganization strategy works in concert with a fine-tuned model to identify the optimal SQL.
XiYan-SQL framework achieves SOTA performance on the well-known Text-to-SQL benchmarks, including BIRD and Spider.

Our work highlights the power of an ensemble approach that synergizes the strengths of both specialized fine-tuned models and large-scale ICL models. 
This combination effectively mitigates the diversity limitations inherent in any single generation method. 
Looking forward, our framework offers several exciting avenues for future research. 
We plan to enhance the intrinsic reasoning capabilities of our specialized models, potentially by integrating more sophisticated fine-tuning strategies like self-reflection or constitutional AI principles. Furthermore, we aim to expand the scope of XiYan-SQL beyond pure generation, exploring its potential as a unified ``all-in-one" model capable of handling a broader range of SQL-related tasks such as schema linking and query optimization.



\bibliographystyle{IEEEtran} 
\bibliography{ref} 

\end{document}